\documentclass[runningheads]{llncs}
\usepackage{amsmath} 
\usepackage{algorithm}
\usepackage{algpseudocode}
\usepackage[dvipsnames]{xcolor}
\usepackage[T1]{fontenc}
\usepackage{graphicx}
\begin{document}
\title{Enhancing Internet of Things Security through Self-Supervised
Graph Neural Networks
}
%
\titlerunning{Enhancing Internet of Thing Security}
%
\author{Safa Ben Atitallah\inst{1,2}\orcidID{0000-0003-0796-3507} %
\and
Maha Driss\inst{1,2} \orcidID{0000-0001-8236-8746}
\and
Wadii Boulila\inst{1,2}\orcidID{0000-0003-2133-0757}
\and 
Anis Koubaa\inst{1}\orcidID{0000-0003-3787-7423}
}
\authorrunning{S. Ben Atitallah et al.}
%
\institute{ RIOTU Lab, CCIS, Prince Sultan University, Riyadh 12435, Saudi Arabia \and RIADI Laboratory, University of Manouba, Manouba 2010, Tunisia
}

\maketitle              
\begin{abstract}
With the rapid rise of the Internet of Things (IoT), ensuring the security of IoT devices has become essential. One of the primary challenges in this field is that new types of attacks often have significantly fewer samples than more common attacks, leading to unbalanced datasets. Existing research on detecting intrusions in these unbalanced labeled datasets primarily employs Convolutional Neural Networks (CNNs) or conventional Machine Learning (ML) models, which result in incomplete detection, especially for new attacks.
To handle these challenges, we suggest a new approach to IoT intrusion detection using Self-Supervised Learning (SSL) with a Markov Graph Convolutional Network (MarkovGCN).
Graph learning excels at modeling complex relationships within data, while SSL mitigates the issue of limited labeled data for emerging attacks.
Our approach leverages the inherent structure of IoT networks to pre-train a GCN, which is then fine-tuned for the intrusion detection task. 
The integration of Markov chains in GCN uncovers network structures and enriches node and edge features with contextual information. 
Experimental results demonstrate that our approach significantly improves detection accuracy and robustness compared to conventional supervised learning methods. Using the EdgeIIoT-set dataset, we attained an accuracy of 98.68\%, a precision of 98.18 \%, a recall of 98.35\%, and an F1-Score of 98.40\%.

\end{abstract}
\section{Introduction}
\label{sec:introduction}
The fast rise of Internet of Things (IoT) devices has revolutionized different fields, from smart homes and healthcare to industrial automation and smart cities \cite{atitallah2020leveraging,heidari2023internet}. However, this increasing number of interconnected devices has also increased security vulnerabilities, making IoT networks a potential target for cyber-attacks. Effective intrusion detection systems are required for identifying and mitigating these threats to ensure the integrity, confidentiality, and availability of IoT systems \cite{moustafa2023explainable}. 

Current methods for intrusion detection in IoT networks face significant limitations, mainly when working with a large number of classes in a dataset, which often leads to decreased classification performance. Conventional approaches primarily rely on Machine Learning (ML) algorithms \cite{sarhan2022feature} or Convolutional Neural Networks (CNNs) \cite{kan2021novel}. These traditional supervised learning methods require large labeled datasets, often challenging to obtain \cite{tsimenidis2022deep}. While some recent works utilize Graph Neural Networks (GNNs), these approaches typically process only a subset of the data due to computational constraints, which can result in overfitting and suboptimal performance \cite{tsimenidis2022deep}.

To address these challenges, we propose to use the MarkovGCN, an integration of Markov chains with GNNs \cite{rahman2022markovgnn}, enhanced by self-supervised learning (SSL).
GNNs excel at modeling complex relationships within data by representing it as graphs where nodes define entities and edges illustrate connections between them. On the other hand, SSL leverages the data to generate labels and learn from them, which permits the model to capture robust illustrations from unlabeled data. 
By employing the MarkovGCN for IoT intrusion detection, we enhance the model's efficiency and demonstrate significant improvements in managing the complexity and variety of attack classes present in IoT environments.
Combining Markov chains with GNNs generates a series of Markov matrices that capture the graph's dynamic relationships and community structures. This allows the model to understand better and represent the data's inherent structure, leading to improved detection capabilities. 
Furthermore, this integration enhances resistance to outliers by focusing on the overall transition patterns rather than working on single instances.
The generated Markov chain transitions are computationally efficient, which makes the model scalable to large IoT networks. By leveraging the full dataset and the inherent properties of Markov chains, the MarkovGCN is less prone to overfitting, especially in scenarios with limited labeled data.

An essential component of our approach is the use of SSL with edge weight prediction. For the SSL task, we predict the edge weights as the pretext task. Given the node feature matrix $X$ and the aggregated edge weights $W$, we train the model to predict the edge weights, which helps in learning robust node representations. 
This is achieved through a series of steps, including node feature transformation, edge weight prediction using a multi-layer perceptron (MLP), and a mean squared error (MSE) loss function to calculate the variance between the real and forecasted edge weights.
The proposed MarkovGCN was evaluated using the EdgeIIoT-Set dataset. It achieved a good performance with an accuracy of 98.68\% and better handled the intricacies of IoT network traffic compared to the conventional models. 

The following points represent illustrations of the study's main contributions:
\begin{itemize}
    \item This paper proposes an intrusion detection approach for IoT networks founded on a combination of MarkovGCN and SSL. The integration of Markov chains within GCN enhances the robustness and scalability of the model and enables it to effectively capture dynamic relationships and community structures within IoT network data. The use of SSL addresses the challenge of limited labeled data through the use of edge weight prediction as a pretext task.

    \item The proposed approach is validated through the use of a public dataset, the Edge-IIoTSet dataset. This dataset includes 15 distinct categories, with 14 classes representing five intrusion types (DDOS, Info-gathering, Malware, Injection, MITM) and a normal class.

    \item The proposed approach achieves strong performance with an accuracy of 98.68\%. Confusion matrix and ROC analysis results are provided.
    
\end{itemize}

The rest of the paper is structured in the following order: Section 2 provides a review of the related works. Section 3 details the architecture of the proposed MarkovGCN in the context of SSL for intrusion detection. Section 4 presents our experimental methodology and evaluation metrics, discusses the results, and compares them with baseline methods. Finally, Section 5 closes with an outline of results and suggestions for future research.

\section{Related Work}

IoT intrusion detection has been studied using various ML techniques. Works in \cite{shafique2020detecting,saheed2022machine,hamza2023malware} proposed supervised ML-based intrusion detection systems for IoT. Different ML models were employed for analysis, including support vector machine, decision tree, random forest, logistic regression, etc. While these methods provide good results, they struggle with imbalanced datasets, particularly when some classes have only a few samples. 

Different works have been proposed for intrusion detection using CNNs \cite{atitallah2022novel,ben2022effective}. CNNs offer several advantages in this context, such as their power to automatically identify relevant features from raw data and their high accuracy in detecting known attack patterns. However, they also come with disadvantages, including their reliance on large amounts of labeled training data, their susceptibility to adversarial attacks, and their potential problem in capturing complex temporal dependencies inherent in network traffic data.
In \cite{tareq2022analysis} two CNN models, DenseNet and Inception Time, were trained using multi-class classification methods to detect cyber-attacks. Evaluation of three datasets—the ToN-IoT, Edge-IIoTSet, and UNSW2015—demonstrated robust performance across various attack types. 

On the other hand, GNNs have been presented as a breakthrough technique for many domains, including network security \cite{bilot2023graph}. Unlike CNNs, GNNs excel in modeling relational data, making them suitable for capturing complex relationships and dependencies within network graphs. This capability allows GNNs to efficiently detect evolving attack patterns that are not explicitly represented in training data. Different approaches in the literature have been proposed based on the GNN for IoT intrusion detection \cite{zhou2021hierarchical,lo2022graphsage,zhang2022intrusion}. However, challenges, including scalability with large graphs and the need for effective graph representation learning, remain areas of active research in applying GNNs to intrusion detection.

Current intrusion detection systems in IoT networks rely mainly on supervised learning models trained on huge labeled data. However, these approaches struggle with detecting novel attacks and may require frequent updates to accommodate new threats. 
Recent advances in SSL have enabled the use of unlabeled data for effective model training. 
Works in \cite{wang2023robust,nguyen2023ts,ben2024strengthening} investigate the use of SSL for intrusion detection and showcase their potential to alleviate the dependency on huge labeled data and provide good performance. For instance, Wang et al. \cite{wang2021network} introduce a data augmentation strategy to improve feature extraction for intrusion detection systems. Similarly, Nguyen et al. \cite{nguyen2023ts} present a transformer-based SSL approach with a masked context reconstruction module that can capture temporal contexts and enhance anomaly detection resilience.

Our work extends these advancements by integrating SSL with MarkovGCN for IoT intrusion detection. 
Integrating SSL with GNNs enables robust graph representation learning and a comprehensive understanding of contextual IoT network behaviors. This combined approach aims to improve the accuracy and efficiency of intrusion detection systems in dynamic and diverse IoT environments.
  
\section{Proposed Approach}

\begin{figure}
    \centering
    \includegraphics[width=0.95\linewidth]{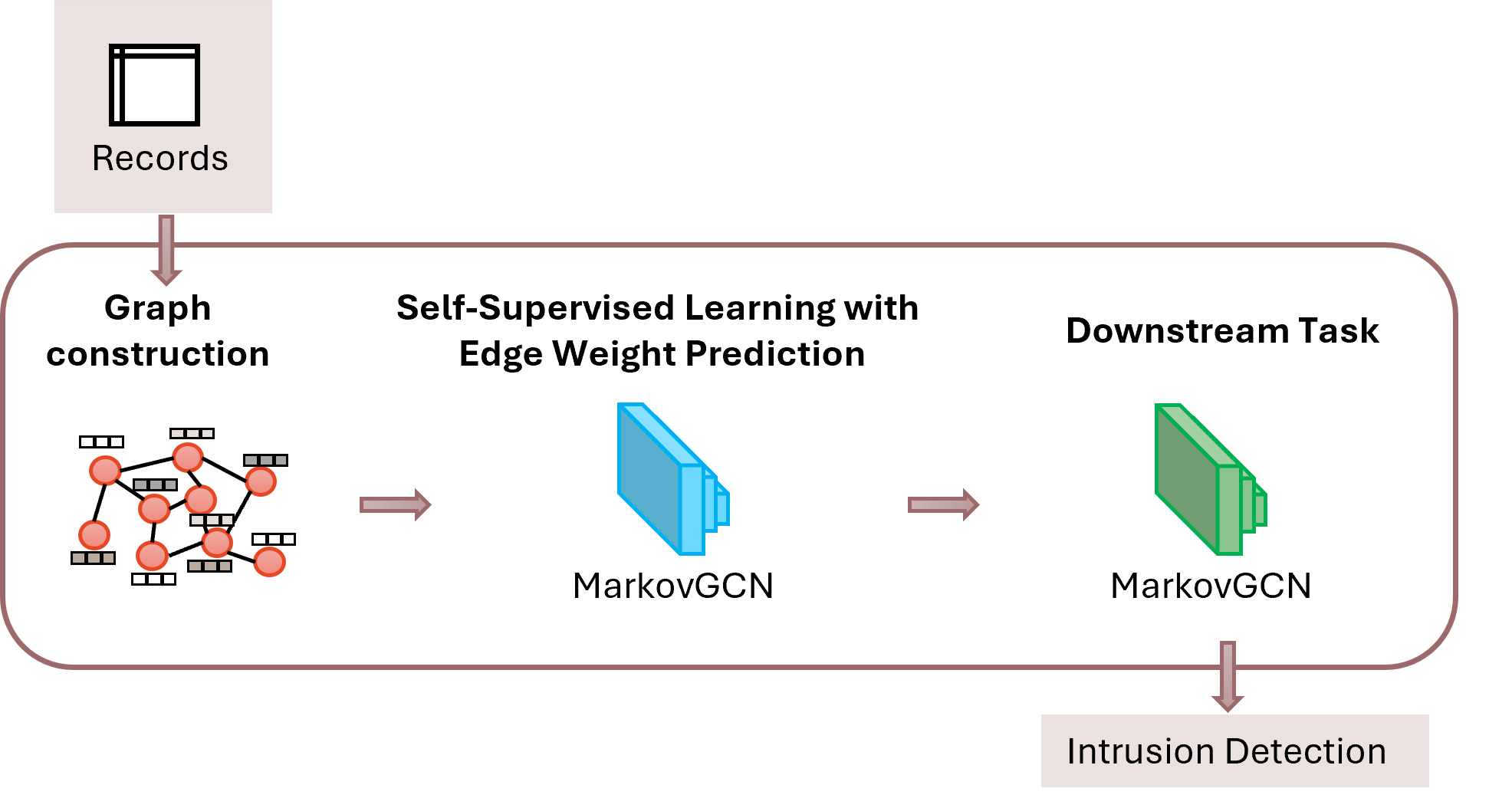}
    \caption{An Overview of the Suggested SSL-Based MarkovGCN for IoT Intrusion Detection}
    \label{fig:approach}
\end{figure}

This section presents the architectural structure of our approach, which is suggested to address challenges encountered in IoT intrusion detection systems. Our objective is to create an SSL-based GCN that can effectively learn and identify IoT intrusions.  
Figure \ref{fig:approach} depicts the structure of the suggested IoT intrusion detection approach. The framework consists of three main stages: graph construction, SSL using edge weight prediction, and a downstream task focused on intrusion detection. The input data is first transformed to a graph from records, where nodes and edges define entities and their relationships. A MarkovGCN model is then applied to perform self-supervised learning to predict edge weights, enhancing the graph's structure. Finally, the learned representations are used in the downstream task for detecting intrusions.

\subsection{Graph Constraction}
In the context of IoT networks, the flow of data and communication is represented as a graph $G=(V,E)$, where $V$ represents the nodes (endpoints), and $E$ is a set of edges. The edges represent the flows between nodes, indicating relationships or connections between them. Each edge connects two nodes and can have associated weights.

\subsection{Self-Supervised Learning with Edge Weight Prediction}

\subsubsection{MarkovGCN Architecture:}

The core of our approach is the MarkovGCN model, designed specifically to capture higher-order dependencies in graph data \cite{rahman2022markovgnn}. MarkovGCN extends traditional GCNs by incorporating Markov chain-based transitions between nodes, thereby enriching node representations. In this network, the transition probabilities derived from the structure of the graph enhance the node feature aggregation process.
The forward propagation in MarkovGCN is defined in the following equation \ref{eq:h}:

\begin{equation}
\label{eq:h}
H^{(l+1)} = \sigma(\tilde{D}^{-\frac{1}{2}} \tilde{A} \tilde{D}^{-\frac{1}{2}} H^{(l)} W^{(l)})
\end{equation}

where \( H^{(l)} \) represents node features at layer \( l \), \( \tilde{A} = A + I \) is the self-looping adjacency matrix, \( \tilde{D} \) is the degree of the matrix of \( \tilde{A} \), \( W^{(l)} \) is the weight matrix of layer \( l \), and \( \sigma \) is the activation function.

\subsubsection{Markov Process Aggregation:}

We begin by applying a Markov process to aggregate information from the graph. Given a graph \( G = (V, E) \) with nodes \( V \) and edges \( E \), the Markov process iteratively updates the edge weights based on the transition probabilities. 
Let \( A \) be the graph's adjacency matrix and \( D \) be the diagonal degree matrix. The transition matrix \( P \) is defined in equation \ref{eq:p}:
\begin{equation}
    \label{eq:p}
    P = D^{-1}A
\end{equation}

In each iteration \( t \), we update the edge weights \( w_{ij}^{(t)} \) using the transition probabilities as presented in equation \ref{eq:w}:
\begin{equation}
    \label{eq:w}
    w_{ij}^{(t+1)} = \frac{P_{ij}^k}{\sum_{j} P_{ij}^k}
\end{equation}

where \( k \) is the inflation parameter controlling the influence of each edge.

\subsubsection{Pretext Task:}
For the SSL task, the prediction of the edge weights is employed as a pretext task. Given the node feature matrix \( X \) and the aggregated edge weights \( W \), we train the model to predict the edge weights using the following steps:

\begin{enumerate}
    \item \textbf{Node Feature Transformation:}

\begin{equation}
    \label{eq:nft}
    H^{(l+1)} = \sigma \left( \sum_{j \in \mathcal{N}(i)} \frac{w_{ij}}{\sqrt{d_i d_j}} H^{(l)} W^{(l)} \right)    
\end{equation}
where \( H^{(l)} \) is layer's feature matrix \( l \), \( W^{(l)} \) is the matrix of trainable weights, \( \mathcal{N}(i) \) is the group of node's neighbors \( i \), \( d_i \) and \( d_j \) represent the degrees of nodes \( i \) and \( j \), and \( \sigma \) is a non-linear activation function.

    \item \textbf{Edge Weight Prediction:}

For each edge \( (i, j) \), we predict the edge weight \( \hat{w}_{ij} \) using a linear transformation of the concatenated node features:
\begin{equation}
    \label{eq:ewp}
    \hat{w}_{ij} = \text{MLP} \left( [H_i \| H_j] \right)    
\end{equation}

where \( [H_i \| H_j] \) denotes the combination of the feature vectors of nodes \( i \) and \( j \), and MLP is a multi-layer perceptron.

    \item \textbf{Loss Function:}

We use MSE to measure the variation between the forecasted edge weights \( \hat{w}_{ij} \) and the true edge weights \( w_{ij} \):
\begin{equation}
    \label{eq:loss}
    \mathcal{L}_{\text{pretext}} = \frac{1}{|E|} \sum_{(i,j) \in E} ( \hat{w}_{ij} - w_{ij} )^2
\end{equation}
\end{enumerate}

\subsection{Downstream Task}

After training the model on the edge weight prediction task, we fine-tune it for the downstream task, intrusion detection. The learned representations from the pretext task serve as a strong initialization for the downstream task.

\begin{enumerate}
    \item \textbf{{Intrusion Detection:}} We apply the GCN for intrusion detection by predicting the class labels \( \hat{y}_i \) for each node \( i \):
\begin{equation}
    \label{eq:ID}
    \hat{y}_i = \text{softmax}(H_i W^{(L)})
\end{equation}

where \( W^{(L)} \) is the trainable weight matrix of the final layer and \( \text{softmax} \) is the softmax activation function.
    \item \textbf{Loss Function:} We employ the Negative Log-Likelihood (NLL) loss for the intrusion detection task:
\begin{equation}
    \label{eq:lf}
    \mathcal{L}_{\text{downstream}} = - \sum_{i \in V_{\text{train}}} y_i \log(\hat{y}_i)    
\end{equation}

where \( y_i \) presents the true label of node \( i \) and \( V_{\text{train}} \) is the set of training nodes.

\end{enumerate}

Algorithm \ref{alg} summarizes the procedure of our proposed approach using MarkovGCN and following SSL for intrusion detection in IoT environments.

\begin{algorithm}
\caption{Workflow Algorithm of the Proposed SSL- based MarkovGCN}
\label{alg}
\begin{algorithmic}[1]
\Require Graph dataset \( D \) with node features and edge attributes
\Ensure Trained MarkovGCN model 
\State \textbf{Begin}
\State Load graph dataset \( D \)
    \State Apply Markov process to aggregate edge attributes:
    \State \hspace{1em} \( (markov\_edges, markov\_weights) = \text{markov\_process\_agg}(D, \epsilon, inflate, nlayers) \)
    \State Define train, validation, and test masks.
    \State Initialize MarkovGCN model and optimizer.
    \For{$epoch \in \{1, \ldots, \text{pretext\_epochs}\}$}
        \State Predict edge weights using the model.
        \State Compute and backpropagate MSE loss.
        \State Update model parameters.
    \EndFor
    \State Fine-tune the model for node classification:
    \For{$epoch \in \{1, \ldots, \text{fine\_tune\_epochs}\}$}
        \State Perform node classification using the model.
        \State Compute and backpropagate NLL loss.
        \State Update model parameters.
    \EndFor
\end{algorithmic}
\end{algorithm}

\section{Experiments}
The following details the experimental methodology employed in this study. We begin by describing the dataset used, outlines the experimental setup, presents the results obtained, compares our findings with related work, and finally, provides a comprehensive discussion.

\subsection{Dataset}

\begin{figure}[h]
    \centering
    \includegraphics[width=0.8\textwidth]{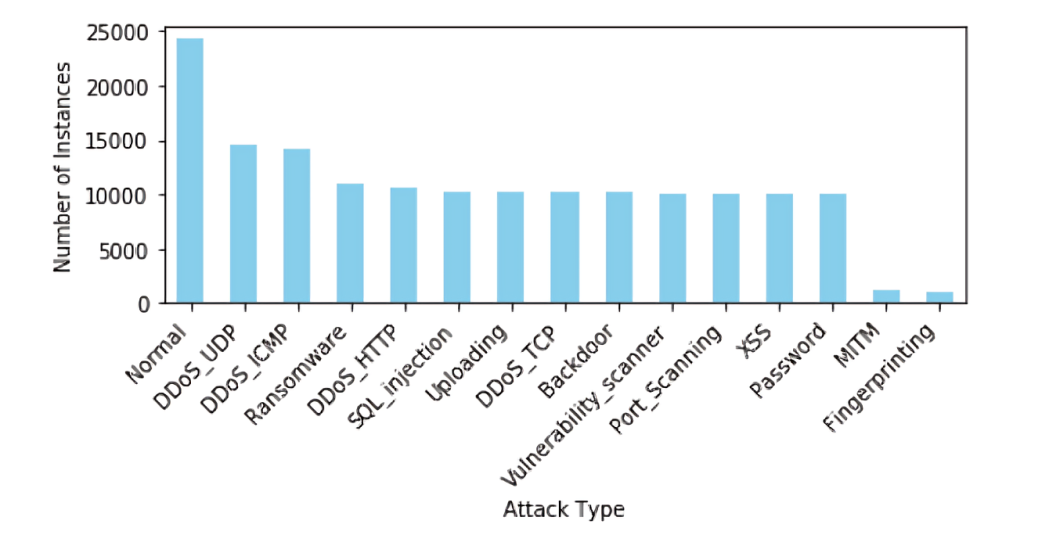}
    \caption{ The distribution of Edge-IIoTSet dataset over the 15 classes}
    \label{fig:enter-label}
\end{figure}

Recently, in 2022, the Edge-IIoTSet dataset was presented by Ferrag \textit{et al.} \cite{ferrag2022edge} as a cyber security dataset. It encompasses various devices, sensors, protocols, and configurations from cloud and edge environments. This dataset consists of 14 attack types linked to IoT connection protocols. These attack types are organized in five main attack groups including: Distributed Denial of Service (DDoS), Injection attacks, Man-in-the-Middle (MITM) attacks, Malware, and Information Gathering, along with Normal traffic patterns, all characterized by 62 features.
The bar chart in Fig. \ref{fig:enter-label} illustrates the distribution of attack types in the Edge-IIoTSet dataset across 15 different classes.
This distribution highlights a notable imbalance in the dataset, with some classes being significantly underrepresented, particularly for the MITM and Fingerprinting.

\subsection{Experimental Settings}
The experiments were performed using a machine featured with an Intel(R) Core(TM) i7-8565U CPU @ 1.80GHz (up to 1.99 GHz), 16 GB of RAM, running Windows 11, and supported by an NVIDIA GeForce MX graphics card. The evaluation metrics employed to assess the suggested approach include accuracy, precision, recall, and F1-score. In addition, we have used the Receiver Operating Characteristic (ROC) curve to assess and visualize the performance of our approach in classifying the different attack types. It plots the true positive rate against the false positive rate.

\subsection{Experimental Results}

\begin{figure*}[h]
    \centering
    \includegraphics[width= 0.9\textwidth] {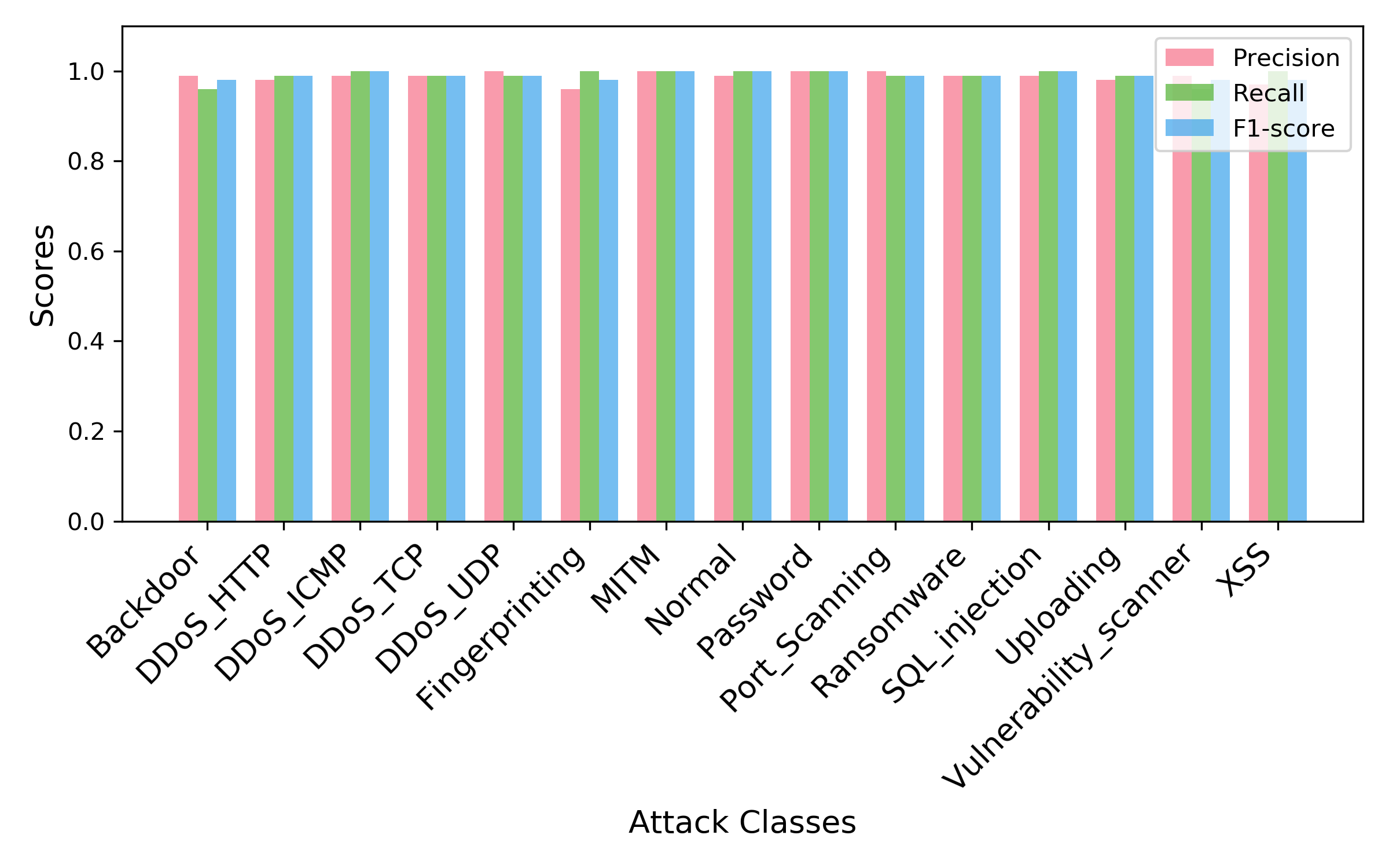}
    \caption{Performance Results: Precision, Recall, and F1-score by Attack Class}
    \label{fig:barchart}
\end{figure*}

Using the EdgeIIoTSet dataset, the suggested approach attains an accuracy of 98.68\%.
The bar chart in Fig. \ref{fig:barchart} shows the precision, recall, and F1-score metrics for the 15 classes included in the dataset. Each bar represents a different attack class in addition to the normal class.  

Most attack classes, including the different types of DDoS, the Normal class, and Ransomware, exhibit high precision, recall, and F1-scores, generally close to 100\%. This denotes the model's robust performance in identifying these attack classes.
The results remain commendable despite the MITM and Fingerprinting classes having relatively fewer instances. This effectiveness can be attributed to the MarkovGCN and SSL deployment. Using the MarkovGCN improves the model's capacity to extract intricate dependences from the structure of the, while SSL enables the model to get robust representations from the data without relying heavily on labeled instances. These methodologies collectively contribute to the model's strong performance through all classes, including those with fewer instances, ensuring good precision, recall, and F1-scores.

\begin{figure*}[h]
    \centering
    \includegraphics[width= 0.98\textwidth]{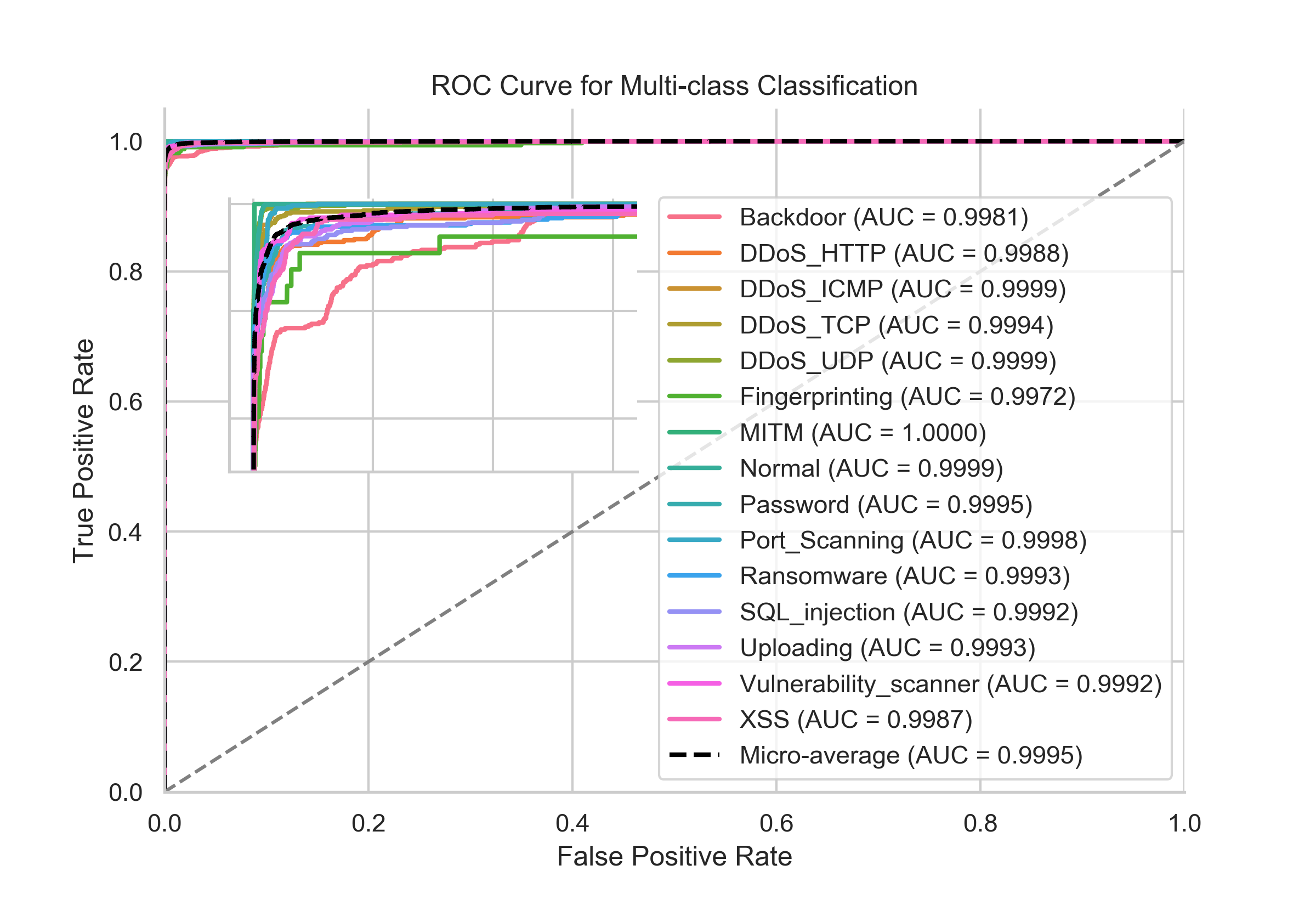}
    \caption{The ROC Curve for Multi-class Classification Using The Proposed Approach on The Edge-IIoTSet dataset. }
    \label{fig:roc}
\end{figure*}

The ROC curve depicted in Fig. \ref{fig:roc} describes the performance of a multi-class classification model across various dataset classes. Each curve corresponds to a different class, with colors chosen for clarity and differentiation. The micro-average curve (dashed line) summarizes overall performance across all classes and reaches a value of 99.95\% of area under the curve (AUC) metric. A higher AUC indicates better discriminatory capability of the model across different attacks.

\subsection{Comparison with Related Works}
To assess the presented approach, we compare our proposed methodology with the work of Ferrag et al. \cite{ferrag2022edge}, who suggest the dataset and employed a straightforward Deep Neural Network (DNN) following the supervised learning. As shown in Table \ref{tab:comp}, our method significantly surpasses this study in terms of performance, confirming the effectiveness of our MarkovGNN model combined with SSL.
\begin{table}[h]
\centering
\caption{Comparison of the proposed approach results}
\label{tab:comp}
\begin{tabular}{lcccccc}
\hline
Model & Type of learning & Architecture & Accuracy & Precision & Recall & F1-score  \\ \hline

Ferrag et al. \cite{ferrag2022edge}& Supervised learning & DNN &  94.67\%  & - & - & -  \\ \hline
Proposed approach  & SSL  & MarkovGNN &   98.68\%     &   98.18\%    &   98.35\%    &  98.40\% \\ \hline
\end{tabular}
\end{table}

\subsection{Discussion}
Our results indicate that MarkovGCN effectively captures the structure of IoT networks, leading to improved intrusion detection. The use of a Markov GNN in a SSL approach offers several contributions and advantages:

\begin{itemize}
    \item  MarkovGCN can effectively model the transitions between nodes. In a SSL context, where the goal is to learn representations without labeled data, capturing these structural transitions helps the model learn more accurate and meaningful graph representations. 
    \item MarkovGCN can be integrated with self-supervised tasks that predict the next step or transition in a sequence. For instance, we enhanced the model's power to extract intricate dependencies within the graph structure by predicting the weight of edges based on the learned Markov transitions. 
    \item In SSL, leveraging the inherent structure and transitions in the data lead to more data-efficient learning. MarkovGCN reduces the labeled data needed for downstream tasks by efficiently using the sequential data.

    \item Markov processes are more robust to noise in the data. Data can be noisy or incomplete in many real-world scenarios, especially in IoT. A MarkovGCN helps in smoothing out this noise by focusing on the underlying transition dynamics rather than just the static node features.
    
\end{itemize}

\section{Conclusion}
IoT networks' growing complexity and scale demand advanced and effective intrusion detection systems. Traditional methods often fall short due to IoT environments' dynamic and heterogeneous nature. In this paper, we present a new approach for IoT intrusion detection using SSL with MarkovGCN. Our method leverages the structural properties of IoT networks to enhance feature learning and improve detection accuracy. 
Using the EdgeIIoT-Set, the results reveal the potential of graph SSL for IoT security, opening the door for reliable and scalable intrusion detection systems.
In future work, we aim to further enhance our model by exploring advanced graph augmentation techniques and incorporating federated learning to improve privacy and efficiency \cite{siddique2024sustainable}. Additionally, we seek to extend our evaluation to more diverse IoT datasets and real-world scenarios, investigate the integration of our approach with other deep learning models to enhance the detection of new cyber-attacks, and build a microservices-based architecture to ensure interoperability, scalability, and resilience of the proposed approach \cite{atitallah2023revolutionizing,atitallah2022microservices} .

\section{Aknowledgement}
The authors thank Prince Sultan University for supporting the conference registration fees financially.

%
%
\bibliographystyle{IEEEtran} 

\end{document}